\pdfoutput=1

\documentclass[11pt]{article}

\usepackage{graphicx}

\usepackage[final]{acl}
\usepackage{hyperref}
\usepackage{booktabs} 

\usepackage{times}
\usepackage{latexsym}
\usepackage{amsmath}
\usepackage{float} 
\usepackage{array}
\newcolumntype{P}[1]{>{\centering\arraybackslash}p{#1}}

\usepackage[T1]{fontenc}

\usepackage[utf8]{inputenc}

\usepackage{microtype}

\usepackage{inconsolata}

\title{BAMO at SemEval-2024 Task 9: BRAINTEASER: A Novel Task Defying Common Sense}

\author{Baktash Ansari, Mohammadmostafa Rostamkhani, Sauleh Eetemadi \\
  Iran University of Science and Technology \\
  \small{\texttt{\{baktash\_ansari, mo\_rostamkhani97\}@comp.iust.ac.ir}, \texttt{sauleh@iust.ac.ir}}}

\begin{document}
\maketitle
\begin{abstract}
This paper outlines our approach to SemEval 2024 Task 9, BRAINTEASER: A Novel Task Defying Common Sense. The task aims to evaluate the ability of language models to think creatively. The dataset comprises multi-choice questions that challenge models to think 'outside of the box'. We fine-tune 2 models, BERT and RoBERTa Large. Next, we employ a Chain of Thought (CoT) zero-shot prompting approach with 6 large language models, such as GPT-3.5, Mixtral, and Llama2. Finally, we utilize ReConcile, a technique that employs a ‘round table conference’ approach with multiple agents for zero-shot learning, to generate consensus answers among 3 selected language models. Our best method achieves an overall accuracy of 85 percent on the sentence puzzles subtask.

\end{abstract}

\section{Introduction}

Evaluation methods in the NLP community predominantly emphasize Vertical thinking, characterized by sequential, analytical processes based on rationality, logic, and rules. However, SemEval-2024 Task 9, BRAINTEASER \cite{jiang-ilievski-ma:2024:SemEval2024}, which is based on the original BRAINTEASER dataset \cite{jiang-etal-2023-brainteaser}, aims to introduce a task that promotes lateral thinking (or ``thinking outside the box’'), a divergent and creative process involving the exploration of new perspectives when addressing problems. The BRAINTEASER QA task consists of two subtasks for the English language: Sentence Puzzles and Word Puzzles. This task is designed to challenge the common sense reasoning capabilities of NLP models and stimulate the development of models that can think laterally.

\begin{itemize}
    \item \textbf{Sentence Puzzles}: Sentence-type brain teaser where the puzzle-defying commonsense is centered on sentence snippets.
    \item \textbf{Word Puzzles}: Word-type brain teaser where the answer violates the default meaning of the word and focuses on the letter composition of the target question.
\end{itemize}

\begin{figure}[htbp]
  \centering
  \includegraphics[width=0.9\linewidth]{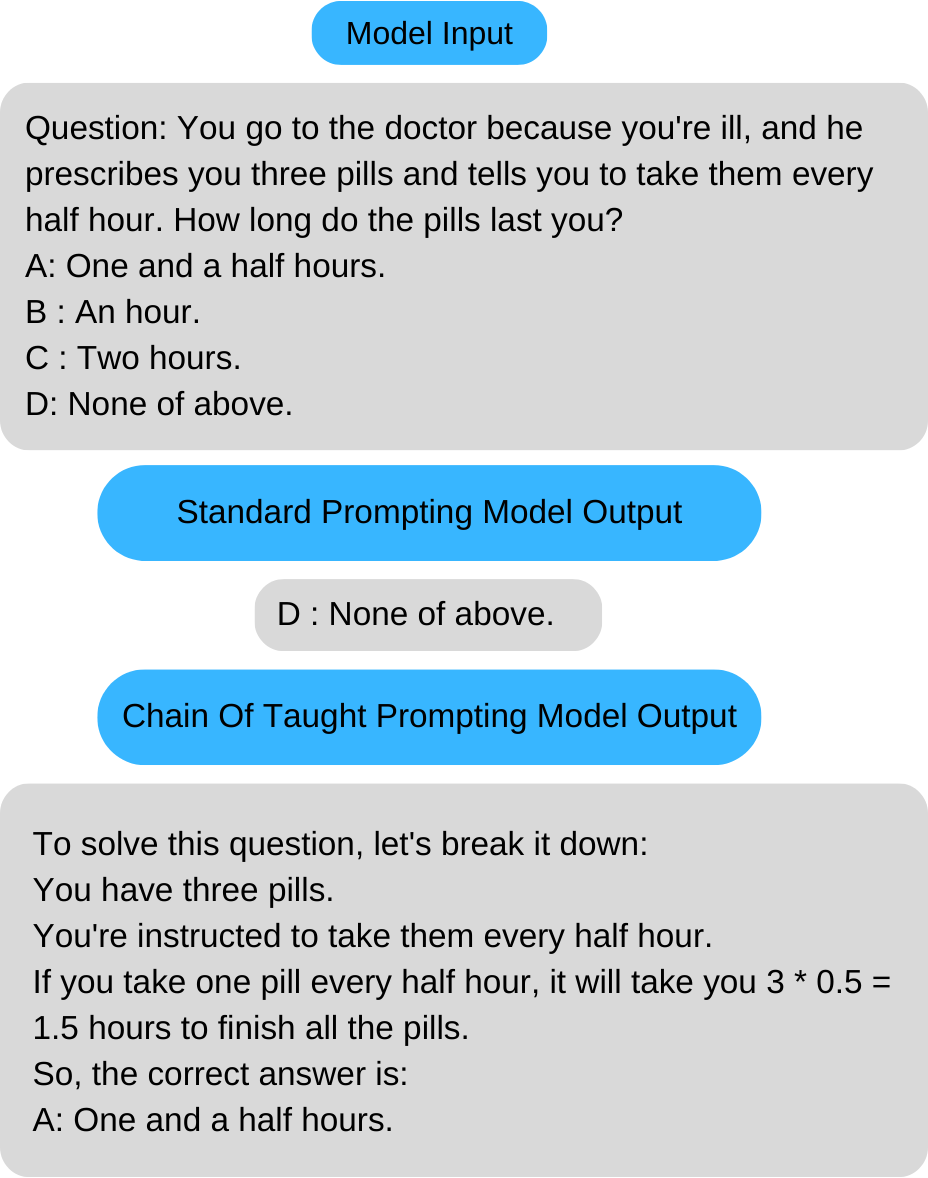}
  \caption{Chain Of Thought Prompting (GPT3.5)}
  \label{fig:fig1}
\end{figure}

We generate baselines for two attention-based models, BERT \cite{devlin-etal-2019-bert} and RoBERTa-Large \cite{liu2019roberta}, as selected in the task paper, to solve these types of multiple-choice problems. Then we fine-tune them with the same configs. After achieving some accuracy through fine-tuning, we explore zero-shot prompting with various large language models (LLMs). To further improve results, zero-shot prompting is conducted using a Chain of Thought technique \cite{wei2023chainofthought}. As illustrated in Figure \ref{fig:fig1}, we compel the model to analyze and provide step-by-step reasoning for its answer instead of simply providing a correct option alone. This approach helps the model focus more on details and answer questions with fewer errors.

\begin{figure*}[htbp]
  \centering
  \includegraphics[width=0.6\textwidth]{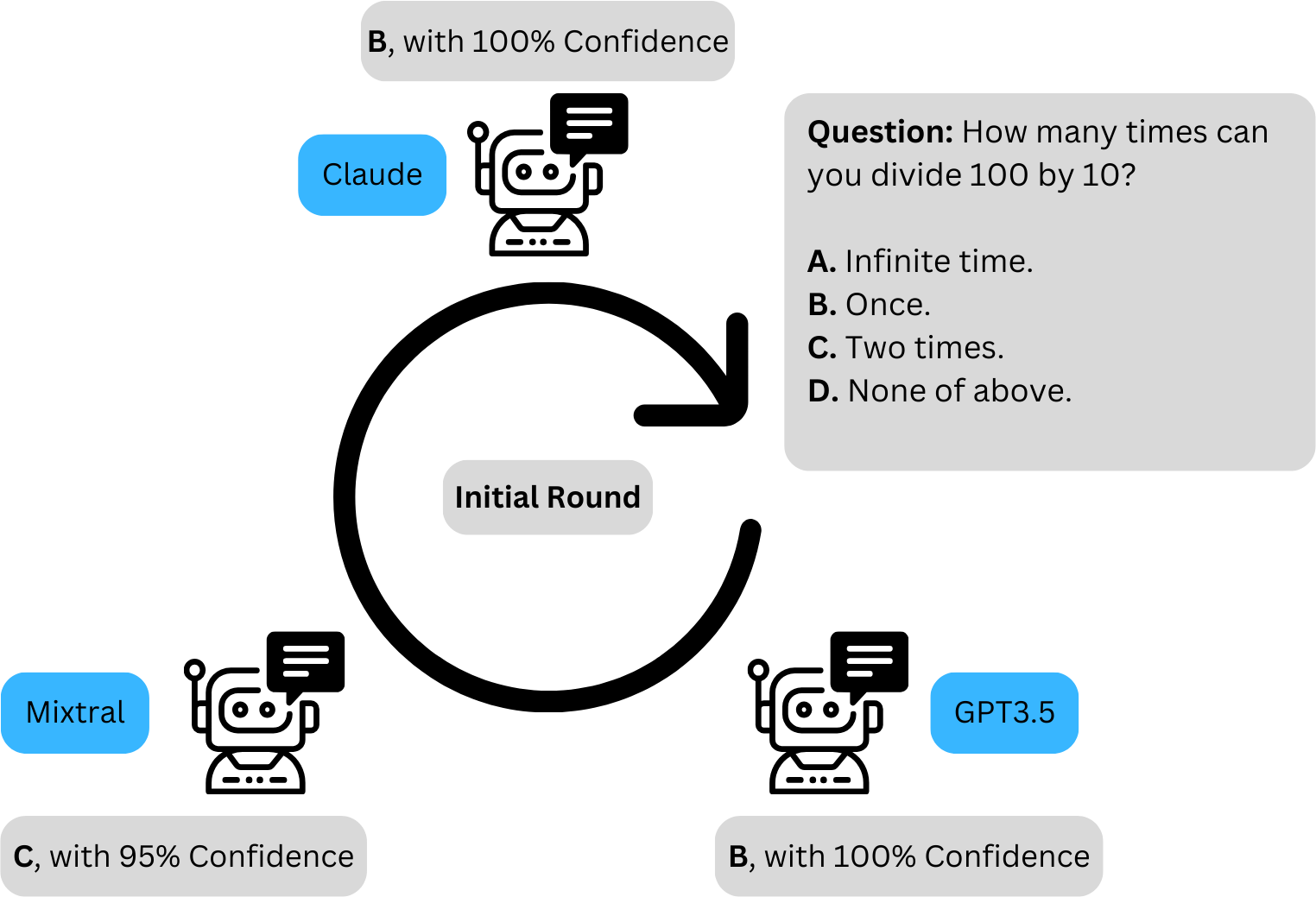}
  \caption{An Illustration of RECONCILE for Initial Round}
  \label{fig:fig2}
\end{figure*}

\newpage

Communication among multiple agents is fundamentally important in complex decision-making processes. Therefore, as our main strategy, we use the ReConcile technique \cite{chen2023reconcile}, which generates answers by gathering the consensus of multiple models based on their confidence levels, to address these problems. To make this technique compatible with this type of dataset, we extend the application of ReConcile from yes/no questions to the BRAINTEASER questions. In this system, we apply the concept of a society of minds to multiple agents. For round table decision-making, we execute three phases: Initial Response Generation, Multi-Round Discussion, and Final Answer Generation on three language learning models: Mixtral8x7b \cite{jiang2024mixtral}, Claude \footnote{Available at \url{https://claude.ai/} .}, and GPT3.5 \footnote{Available at \url{https://openai.com/} .}. As illustrated in Figure \ref{fig:fig2}, in each phase, we generate specific prompts (see Appendix \ref{sec:prompts}) for models to answer the question, along with their confidence level between 0 and 1. We then use their answers for the next round and derive an overall consensus answer for each round. The method that gave us our best result achieved a rank of 11 out of 33 for the sentence puzzles in the task leaderboard. Further details of our implementations are available through our GitHub repository. \footnote{{\href{https://github.com/baktash81/SemEval_2024_BRAINTEASER}{Github Repository}}}.

\section{Background}

\subsection{Related Works}
The exploration of reasoning abilities in large language models, lateral thinking, and common sense reasoning has been the focus of several studies in recent years. The BRAINTEASER is a novel task in this context, requiring a unique blend of these capabilities. In the paper \cite{zhang2022empirical}, foundational insights into the use of knowledge graphs for self-supervision in common sense reasoning tasks are provided. This work is particularly relevant as it demonstrates how external knowledge can significantly improve an LLM’s ability to solve reasoning problems. Furthering the discussion on reasoning, LatEval \cite{huang2024lateval} introduces an evaluation benchmark for LLMs based on lateral thinking puzzles. This process is similar to the method in solving the BRAINTEASER, which encourages models to elaborate on their reasoning, leading to improved performance. The paper RiddleSense \cite{lin2021riddlesense} offers an analysis of how LLMs handle riddle questions that require linguistic creativity and common sense knowledge. This is directly linked to our approaches, where the ability of an LLM to use analogous examples influences its problem-solving effectiveness.  Also, \cite{dou2022zeroshot} investigates the inherent knowledge within language models and their ability to function in zero-shot common sense question answering tasks. The integration of external knowledge is particularly relevant to tasks requiring common sense and real-world information, underscoring the importance of knowledge retrieval in the context of the BRAINTEASER. MVP-Tuning \cite{huang-etal-2023-mvp} introduces a novel approach to knowledge retrieval using prompt tuning. This aligns with our aim in solving BRAINTEASER to evaluate the capacity of LLMs to handle questions that necessitate an understanding of the world as humans perceive it. Lastly, ReConcile \cite{chen2023reconcile} and \cite{liang2023encouraging} both discuss the use of multiple LLMs to improve reasoning capabilities. These works highlight the potential of using a diverse set of models to solve complex tasks like the BRAINTEASER, further enriching the strategies for solving such tasks with AI systems.

\subsection{Datasets}
The organizers provide datasets for one language: English. As mentioned previously, the dataset consists of two categories: Sentence Puzzles and Word Puzzles. The task providers construct reconstruction versions of the original data in two parallel ways: Semantic Reconstruction and Context Reconstruction. This is done to ensure that the task evaluates lateral thinking ability rather than mere memorization. The Semantic variant reformulates the initial question while preserving its answer, whereas the Context variant retains the misleading commonsense assumption as is and modifies the question and its answer to fit a different situational context.

The dataset is split into two parts for the evaluation phase: train and test sets for each category. In the sentence puzzles category, the train set comprises 507-row samples, while the test set consists of 120-row samples. Similarly, in the word puzzles category, the train set contains 396-row samples, and the test set has 96-row samples. Each sample includes a question with its corresponding answer and three distractors. We utilize both word and sentence puzzle datasets during the training phase, but only the sentence puzzles dataset is used for zero-shot phases.

\subsection{Evaluation Metrics} 
The accuracy metric is employed for evaluation as described in the task paper. Performance evaluation is conducted using two accuracy metrics: Instance-based Accuracy and Group-based Accuracy for Original, Semantic, and Context questions.

\section{System overview}

\subsection{Preprocessing}

In the fine-tuning phase, we employ two transformer-based models, BERT and RoBERTa, for multiple choice tasks. Both models are pre-trained on large text corpora. The input to these models is a sequence constructed by concatenating the question with each choice, separated by special tokens. This process is facilitated by the models' tokenizers, which convert the text into a format that the models can understand. 
For a given question "Q", and choices, the input to the models would be: 

\begin{equation}
\text{Input}_i = [\text{CLS}] \text{ Q } [\text{SEP}] \text{Choice}_i [\text{SEP}]
\label{eq:eq0}
\end{equation}

where "i" represents the index of the choice. Each sequence represents a different choice, and the models' understanding of the context and the choices allows them to predict the correct answer. The fine-tuning process adapts the models to this specific task, optimizing their parameters to minimize the difference between the predicted and actual answers.

\subsection{Model Training}
We utilize BERT-Base and RoBERTa-Large models, along with their respective tokenizers, for word embeddings in the multiple choice task, sourced from the Hugging Face library. These models are finetuned using the Hugging Face trainer. Initially, we load the models and establish a baseline on the test set for both sentence and word puzzles. Subsequently, each model undergoes finetuning with the default Hugging Face Cross-Entropy loss function for classification, and overall accuracy is computed as the metric using the same hyper-parameter configurations as detailed in Table \ref{tab:hyperparameters}.

\begin{table}[htbp]
    \centering
    
    \begin{tabular}{ll}
        \toprule
        \textbf{Hyperparameter} & \textbf{Value} \\
        \midrule
        Learning Rate & $1 \times 10^{-5}$ \\
        Optimizer & Adam \\
        $\beta_1, \beta_2$ & $0.9, 0.999$ \\
        Weight Decay & $0.01$ \\
        Batch Size & $1$ \\
        Loss Function & Cross-Entropy \\
        Logging Steps & $100$ \\
        Evaluation Metric & Accuracy \\
        Global Seed & $255$ \\
        \bottomrule
    \end{tabular}
    \caption{Hyperparameters Configuration}
    \label{tab:hyperparameters}
\end{table}

\subsection{Chain of Thought Prompting}
We use 6 LLMs for CoT zero-shot: Mixtral(8x7b), Claude, GPT3.5, Llama-2-70b \cite{touvron2023llama}, OpenChat \cite{wang2023openchat}, and Microsoft Copilot \footnote{Available at \url{https://copilot.microsoft.com/}}. We use graphical web page interfaces for Claude, GPT3.5(chatGPT), and Microsoft Copilot (precise mode). We use the same prompt as the ReConcile initial round. Microsoft Copilot gives the best performance of this section.

\subsection{ReConcile Round Table}
Models make mistakes in one or more types of questions and cannot provide the correct answer on the first attempt. We need to ask them to pay attention to certain parts of the question or give hints to the model so it can provide the correct answer. To ensure human involvement is minimized and models can help each other, we have employed the ReConcile method. Using this approach, each model complements the other. The process of this system is as follows: 

\begin{itemize}
    \item \textbf{Initial Response Generation}: First, using an initial prompt, we ask each model to provide the answer to the question, provide a reason for the answer, and declare a confidence level between 0 and 1.
    \item \textbf{Multi-Round Discussion}: We give the responses, reasoning, and confidence levels of the three models, along with the initial prompt, as input to the models once again. This enables them to consider both the context of the question and the responses of the three models when making a selection.
    \item \textbf{Final Answer Generation.} In this stage, we initialize a weight for each of the 4 options of the question, and these weights are summed up with the confidence level of each model. Finally, the option with the highest weight is chosen as the correct option.
\end{itemize}

Let's denote the confidence of the model \(m_i\) for its selected choice \(c_j\) as \(conf(m_i, c_j)\). Then, the total confidence of each choice can be calculated as:\\
\begin{equation}
    \text{TC}(c_j) = \sum_{i=1}^{3} \text{conf}(m_i, c_j)
    \label{eq:eq1}
\end{equation}

Where the sum is over all models that selected choice \(c_j\). Finally, the choice with the highest total confidence is selected as the correct choice:\\
\begin{equation}
    \text{c}_{\text{correct}} = \arg\max_{c_j \in C} TC(c_j)
    \label{eq:eq2}
\end{equation}

This means that the correct choice is the one that maximizes the total confidence over all choices.\\

The notable point is that this method should be implemented by models that roughly have equal performance to grow together after several rounds. If a model has much lower performance compared to other models, its reasoning and confidence level may negatively affect others. For this reason, we performed this task on three models: Mixtral8x7b, GPT3.5, and Claude, which have almost similar accuracy in the initial round. This iterative process can be continued until all models reach a consensus and all agree on a specific option for the questions. We repeated this process for two discussion rounds.

\begin{table*}[ht]
\centering
\resizebox{\linewidth}{!}{
\begin{tabular}{|c|c|c|c|c|c|c|c|}
\hline
Model & Type & S\_ori & S\_sem & S\_con & S\_ori\_sem & S\_ori\_sem\_con & S\_overall \\ \hline
BERT Base & Baseline & 0.400 & 0.450 & 0.325 & 0.350 & 0.175 & 0.391 \\ \hline
RoBERTa Large & Baseline & 0.250 & 0.175 & 0.275 & 0.175 & 0.050 & 0.233 \\ \hline
BERT Base & Finetune & 0.725 & 0.750 & 0.650 & 0.725 & 0.575 & 0.708 \\ \hline
RoBERTa Large & Finetune & \textbf{0.800} & \textbf{0.775} & \textbf{0.725} & \textbf{0.775} & \textbf{0.700} & \textbf{0.766} \\ \hline
\end{tabular}}
\caption{Models' Performance on Sentence Puzzles}
\label{tab:finetune_s}
\end{table*}

\begin{table*}[ht]
\centering
\resizebox{\linewidth}{!}{
\begin{tabular}{|c|c|c|c|c|c|c|c|}
\hline
Model & Type & W\_ori & W\_sem & W\_con & W\_ori\_sem & W\_ori\_sem\_con & W\_overall \\ \hline
BERT Base & Baseline & 0.562 & 0.343 & 0.375 & 0.281 & 0.093 & 0.427 \\ \hline
RoBERTa Large & Baseline & 0.250 & 0.281 & 0.343 & 0.218 & 0.093 & 0.291 \\ \hline
BERT Base & Finetune & \textbf{0.687} & 0.656 & 0.468 & 0.625 & 0.375 & 0.604 \\ \hline
RoBERTa Large & Finetune & \textbf{0.687} & \textbf{0.687} & \textbf{0.562} & \textbf{0.656} & \textbf{0.468} & \textbf{0.645} \\ \hline
\end{tabular}}
\caption{Models' Performance on Word Puzzles}
\label{tab:finetune_w}
\end{table*}

\begin{table*}[ht]
\centering
\resizebox{\linewidth}{!}{
\begin{tabular}{|c|c|c|c|c|c|c|c|}
\hline
Model & S\_ori & S\_sem & S\_con & S\_ori\_sem & S\_ori\_sem\_con & S\_overall \\ \hline
Open Chat  & 0.500 & 0.500 & 0.525 & 0.375 & 0.300 & 0.508 \\ \hline
Llama-2-70b  & 0.600 & 0.625 & 0.550 & 0.550 & 0.400 & 0.591 \\ \hline
Mixtral-8x7b  & 0.750 & 0.700 & 0.650 & 0.650 & 0.525 & 0.700 \\ \hline
GPT3.5  & 0.750 & \textbf{0.775} & 0.625 & 0.650 & 0.500 & 0.710 \\ \hline
Claud  & \textbf{0.775} & 0.725 & 0.700 & \textbf{0.725} & 0.625 & \textbf{0.730} \\ \hline
Microsoft Copilot  & \textbf{0.925} & \textbf{0.900} & \textbf{0.775} & \textbf{0.875} & \textbf{0.750} & \textbf{0.860} \\ \hline
\end{tabular}}
\caption{LLM's Zero-Shot Performance on Sentence Puzzles}
\label{tab:zeroshot_s}
\end{table*}

\begin{table*}[ht]
\centering
\small
\resizebox{\linewidth}{!}{
\begin{tabular}{|c|c|c|c|c|c|c|}
\hline
Model & S\_ori & S\_sem & S\_con & S\_ori\_sem & S\_ori\_sem\_con & S\_overall \\ \hline
\multicolumn{7}{|c|}{\textbf{Initial Round}} \\ \hline
 GPT3.5 & 0.750 & \textbf{0.775} & 0.625 & 0.650 & 0.500 & 0.710 \\ \hline
 Claude & \textbf{0.775} & 0.725 & \textbf{0.700} & \textbf{0.725} & 0.625 & \textbf{0.730} \\ \hline
 Mixtral-8x7b & 0.750 & 0.700 & 0.650 & 0.650 & 0.525 & 0.700 \\ \hline
 Consensus & \textbf{0.775} & \textbf{0.750} & \textbf{0.700} & \textbf{0.675} & 0.575 & 0.740 \\ \hline
\multicolumn{7}{|c|}{\textbf{Round 1}} \\ \hline
 GPT3.5 & 0.775 & 0.725 & 0.675 & 0.700 & 0.570 & 0.720 \\ \hline
 Claude & \textbf{0.800} & \textbf{0.775} & \textbf{0.725} & \textbf{0.725} & \textbf{0.600} & \textbf{0.760} \\ \hline
 Mixtral-8x7b & 0.700 & 0.725 & 0.625 & 0.625 & 0.450 & 0.680 \\ \hline
 Consensus & \textbf{0.800} & \textbf{0.750} & \textbf{0.700} & \textbf{0.725} & \textbf{0.600} & \textbf{0.750} \\ \hline
\multicolumn{7}{|c|}{\textbf{Round 2}} \\ \hline
 GPT3.5 & \textbf{0.800} & \textbf{0.800} & 0.675 & \textbf{0.750} & \textbf{0.600} & \textbf{0.750} \\ \hline
 Claude & \textbf{0.800} & 0.775 & 0.725 & 0.725 & \textbf{0.600} & \textbf{0.760} \\ \hline
 Mixtral-8x7b & 0.725 & 0.725 & \textbf{0.725} & 0.675 & 0.550 & 0.725 \\ \hline
 Consensus & \textbf{0.775} & \textbf{0.800} & \textbf{0.700} & \textbf{0.725} & \textbf{0.600} & \textbf{0.758} \\ \hline
\end{tabular}}
\caption{Reconcile Results}
\label{tab:reconcile_results}
\end{table*}

\section{Experiments and Results}
\subsection{Experimental Setup}
The training and test sets of the sentence and word puzzle datasets are used with a split of 0.8 and 0.2, respectively for fine-tuning. Additionally, we utilize Google Colab's T4 GPU with the hyperparameters as shown in Table \ref{tab:hyperparameters}. For zero-shot prompting, we use the 120-row test set from the sentence puzzles. 

We leverage HuggingChat \footnote{Available at \url{https://huggingface.co/chat/}} for Mixtral8x7b, OpenChat, and Llama2-70b. Furthermore, we utilize the official web interface of Claude, Microsoft Copilot, and GPT3.5. Results for each model and their corresponding training codes are available in the GitHub repository.

\subsection{Results}

As illustrated in Table \ref{tab:finetune_s} for the sentence puzzles and Table \ref{tab:finetune_w} for the word puzzles, we present the performance of BERT and RoBERTa Large in both their base and fine-tuned versions.\footnote{S(sentence), ori(original), sem(semantic), con(context)} We load the best model based on Overall Accuracy at the end of each training. The best performance is achieved by RoBERTa for both sentence and word puzzles. As illustrated in Appendix \ref{sec:training_logs} For every 100 training steps, we log the overall accuracy for two models.

The submission scores computed by the task organizer for CoT zero-shot are available in Table \ref{tab:zeroshot_s}. Among these LLMs, Microsoft Copilot achieves the best performance. The success of LLMs in responding to these questions depends on the model's ability to recognize that these questions are tricky and that it doesn't need to provide logical reasoning in many cases; the question merely plays with words. Microsoft Copilot understood this phenomenon in many questions. However, it also made mistakes in several questions. For example, consider Appendix \ref{sec:copilot_examples} for an illustration.

For Reconcile, the results of each model in every round, as well as the consensus reached in each round, are presented in Table \ref{tab:reconcile_results}. In the table, we observe that Claude achieves the highest overall accuracy among the models in each round. Nearly every model in the Reconcile system either improves or maintains its best performance in overall accuracy with each round. This suggests that they are all capable of making informed decisions based on the reasoning provided by all agents during the discussion rounds. At the conclusion of round 2, the consensus overall accuracy stands at 0.758, which is 0.3 to 0.5 points higher than the initial round results of all three models. Furthermore, we note that the consensus result improves by approximately 1 percent from the initial round to round 1, and by approximately 0.8 percent from round 1 to round 2. This indicates that after several rounds, the models converge and reach a consensus on the questions. See Appendix \ref{sec:reconcile_convergence}. Also for an example, see Appendix \ref{sec:reconcile_example}

\section{Conclusion}
In this paper, we present our approach to SemEval 2024 Task 9, BRAINTEASER: A Novel Task Defying Common Sense, which challenges models to think creatively beyond conventional reasoning. Through fine-tuning BERT and RoBERTa models, as well as employing zero-shot prompting techniques using various large language models, we achieved notable performance improvements. Particularly, Microsoft Copilot performs the best without being taught beforehand, showing it understands the tricky task really well. Furthermore, our ReConcile Round Table method demonstrates the efficacy of collaborative decision-making among models, leading to a progressive improvement in overall accuracy across multiple rounds of discussion.

\bibliography{main}

\begin{thebibliography}{15}
\expandafter\ifx\csname natexlab\endcsname\relax\def\natexlab#1{#1}\fi

\bibitem[{Chen et~al.(2023)Chen, Saha, and Bansal}]{chen2023reconcile}
Justin Chih-Yao Chen, Swarnadeep Saha, and Mohit Bansal. 2023.
\newblock \href {http://arxiv.org/abs/2309.13007} {Reconcile: Round-table conference improves reasoning via consensus among diverse llms}.

\bibitem[{Devlin et~al.(2019)Devlin, Chang, Lee, and Toutanova}]{devlin-etal-2019-bert}
Jacob Devlin, Ming-Wei Chang, Kenton Lee, and Kristina Toutanova. 2019.
\newblock \href {https://doi.org/10.18653/v1/N19-1423} {{BERT}: Pre-training of deep bidirectional transformers for language understanding}.
\newblock In \emph{Proceedings of the 2019 Conference of the North {A}merican Chapter of the Association for Computational Linguistics: Human Language Technologies, Volume 1 (Long and Short Papers)}, pages 4171--4186, Minneapolis, Minnesota. Association for Computational Linguistics.

\bibitem[{Dou and Peng(2022)}]{dou2022zeroshot}
Zi-Yi Dou and Nanyun Peng. 2022.
\newblock \href {http://arxiv.org/abs/2201.00136} {Zero-shot commonsense question answering with cloze translation and consistency optimization}.

\bibitem[{Huang et~al.(2024)Huang, Ma, Li, Huang, Zou, Zhang, and Zheng}]{huang2024lateval}
Shulin Huang, Shirong Ma, Yinghui Li, Mengzuo Huang, Wuhe Zou, Weidong Zhang, and Hai-Tao Zheng. 2024.
\newblock \href {http://arxiv.org/abs/2308.10855} {Lateval: An interactive llms evaluation benchmark with incomplete information from lateral thinking puzzles}.

\bibitem[{Huang et~al.(2023)Huang, Li, Xu, Zhang, Gan, Zhang, and Wang}]{huang-etal-2023-mvp}
Yongfeng Huang, Yanyang Li, Yichong Xu, Lin Zhang, Ruyi Gan, Jiaxing Zhang, and Liwei Wang. 2023.
\newblock \href {https://doi.org/10.18653/v1/2023.acl-long.750} {{MVP}-tuning: Multi-view knowledge retrieval with prompt tuning for commonsense reasoning}.
\newblock In \emph{Proceedings of the 61st Annual Meeting of the Association for Computational Linguistics (Volume 1: Long Papers)}, pages 13417--13432, Toronto, Canada. Association for Computational Linguistics.

\bibitem[{Jiang et~al.(2024{\natexlab{a}})Jiang, Sablayrolles, Roux, Mensch, Savary, Bamford, Chaplot, de~las Casas, Hanna, Bressand, Lengyel, Bour, Lample, Lavaud, Saulnier, Lachaux, Stock, Subramanian, Yang, Antoniak, Scao, Gervet, Lavril, Wang, Lacroix, and Sayed}]{jiang2024mixtral}
Albert~Q. Jiang, Alexandre Sablayrolles, Antoine Roux, Arthur Mensch, Blanche Savary, Chris Bamford, Devendra~Singh Chaplot, Diego de~las Casas, Emma~Bou Hanna, Florian Bressand, Gianna Lengyel, Guillaume Bour, Guillaume Lample, Lélio~Renard Lavaud, Lucile Saulnier, Marie-Anne Lachaux, Pierre Stock, Sandeep Subramanian, Sophia Yang, Szymon Antoniak, Teven~Le Scao, Théophile Gervet, Thibaut Lavril, Thomas Wang, Timothée Lacroix, and William~El Sayed. 2024{\natexlab{a}}.
\newblock \href {http://arxiv.org/abs/2401.04088} {Mixtral of experts}.

\bibitem[{Jiang et~al.(2024{\natexlab{b}})Jiang, Ilievski, and Ma}]{jiang-ilievski-ma:2024:SemEval2024}
Yifan Jiang, Filip Ilievski, and Kaixin Ma. 2024{\natexlab{b}}.
\newblock \href {https://aclanthology.org/2024.semeval2024-1.271} {Semeval-2024 task 9: Brainteaser: A novel task defying common sense}.
\newblock In \emph{Proceedings of the 18th International Workshop on Semantic Evaluation (SemEval-2024)}, pages 1996--2010, Mexico City, Mexico. Association for Computational Linguistics.

\bibitem[{Jiang et~al.(2023)Jiang, Ilievski, Ma, and Sourati}]{jiang-etal-2023-brainteaser}
Yifan Jiang, Filip Ilievski, Kaixin Ma, and Zhivar Sourati. 2023.
\newblock \href {https://doi.org/10.18653/v1/2023.emnlp-main.885} {{BRAINTEASER}: Lateral thinking puzzles for large language models}.
\newblock In \emph{Proceedings of the 2023 Conference on Empirical Methods in Natural Language Processing}, pages 14317--14332, Singapore. Association for Computational Linguistics.

\bibitem[{Liang et~al.(2023)Liang, He, Jiao, Wang, Wang, Wang, Yang, Tu, and Shi}]{liang2023encouraging}
Tian Liang, Zhiwei He, Wenxiang Jiao, Xing Wang, Yan Wang, Rui Wang, Yujiu Yang, Zhaopeng Tu, and Shuming Shi. 2023.
\newblock \href {http://arxiv.org/abs/2305.19118} {Encouraging divergent thinking in large language models through multi-agent debate}.

\bibitem[{Lin et~al.(2021)Lin, Wu, Yang, Lee, and Ren}]{lin2021riddlesense}
Bill~Yuchen Lin, Ziyi Wu, Yichi Yang, Dong-Ho Lee, and Xiang Ren. 2021.
\newblock \href {http://arxiv.org/abs/2101.00376} {Riddlesense: Reasoning about riddle questions featuring linguistic creativity and commonsense knowledge}.

\bibitem[{Liu et~al.(2019)Liu, Ott, Goyal, Du, Joshi, Chen, Levy, Lewis, Zettlemoyer, and Stoyanov}]{liu2019roberta}
Yinhan Liu, Myle Ott, Naman Goyal, Jingfei Du, Mandar Joshi, Danqi Chen, Omer Levy, Mike Lewis, Luke Zettlemoyer, and Veselin Stoyanov. 2019.
\newblock \href {http://arxiv.org/abs/1907.11692} {Roberta: A robustly optimized bert pretraining approach}.

\bibitem[{Touvron et~al.(2023)Touvron, Martin, Stone, Albert, Almahairi, Babaei, Bashlykov, Batra, Bhargava, Bhosale, Bikel, Blecher, Ferrer, Chen, Cucurull, Esiobu, Fernandes, Fu, Fu, Fuller, Gao, Goswami, Goyal, Hartshorn, Hosseini, Hou, Inan, Kardas, Kerkez, Khabsa, Kloumann, Korenev, Koura, Lachaux, Lavril, Lee, Liskovich, Lu, Mao, Martinet, Mihaylov, Mishra, Molybog, Nie, Poulton, Reizenstein, Rungta, Saladi, Schelten, Silva, Smith, Subramanian, Tan, Tang, Taylor, Williams, Kuan, Xu, Yan, Zarov, Zhang, Fan, Kambadur, Narang, Rodriguez, Stojnic, Edunov, and Scialom}]{touvron2023llama}
Hugo Touvron, Louis Martin, Kevin Stone, Peter Albert, Amjad Almahairi, Yasmine Babaei, Nikolay Bashlykov, Soumya Batra, Prajjwal Bhargava, Shruti Bhosale, Dan Bikel, Lukas Blecher, Cristian~Canton Ferrer, Moya Chen, Guillem Cucurull, David Esiobu, Jude Fernandes, Jeremy Fu, Wenyin Fu, Brian Fuller, Cynthia Gao, Vedanuj Goswami, Naman Goyal, Anthony Hartshorn, Saghar Hosseini, Rui Hou, Hakan Inan, Marcin Kardas, Viktor Kerkez, Madian Khabsa, Isabel Kloumann, Artem Korenev, Punit~Singh Koura, Marie-Anne Lachaux, Thibaut Lavril, Jenya Lee, Diana Liskovich, Yinghai Lu, Yuning Mao, Xavier Martinet, Todor Mihaylov, Pushkar Mishra, Igor Molybog, Yixin Nie, Andrew Poulton, Jeremy Reizenstein, Rashi Rungta, Kalyan Saladi, Alan Schelten, Ruan Silva, Eric~Michael Smith, Ranjan Subramanian, Xiaoqing~Ellen Tan, Binh Tang, Ross Taylor, Adina Williams, Jian~Xiang Kuan, Puxin Xu, Zheng Yan, Iliyan Zarov, Yuchen Zhang, Angela Fan, Melanie Kambadur, Sharan Narang, Aurelien Rodriguez, Robert Stojnic, Sergey Edunov, and Thomas
  Scialom. 2023.
\newblock \href {http://arxiv.org/abs/2307.09288} {Llama 2: Open foundation and fine-tuned chat models}.

\bibitem[{Wang et~al.(2023)Wang, Cheng, Zhan, Li, Song, and Liu}]{wang2023openchat}
Guan Wang, Sijie Cheng, Xianyuan Zhan, Xiangang Li, Sen Song, and Yang Liu. 2023.
\newblock Openchat: Advancing open-source language models with mixed-quality data.
\newblock \emph{arXiv preprint arXiv:2309.11235}.

\bibitem[{Wei et~al.(2023)Wei, Wang, Schuurmans, Bosma, Ichter, Xia, Chi, Le, and Zhou}]{wei2023chainofthought}
Jason Wei, Xuezhi Wang, Dale Schuurmans, Maarten Bosma, Brian Ichter, Fei Xia, Ed~Chi, Quoc Le, and Denny Zhou. 2023.
\newblock \href {http://arxiv.org/abs/2201.11903} {Chain-of-thought prompting elicits reasoning in large language models}.

\bibitem[{Zhang et~al.(2022)Zhang, Ilievski, Ma, Francis, and Oltramari}]{zhang2022empirical}
Jiarui Zhang, Filip Ilievski, Kaixin Ma, Jonathan Francis, and Alessandro Oltramari. 2022.
\newblock \href {http://arxiv.org/abs/2205.10661} {An empirical investigation of commonsense self-supervision with knowledge graphs}.

\end{thebibliography}

\appendix

\section{Training logs}
\label{sec:training_logs}

\begin{figure}[ht]
  \centering
  \includegraphics[width=1\linewidth]{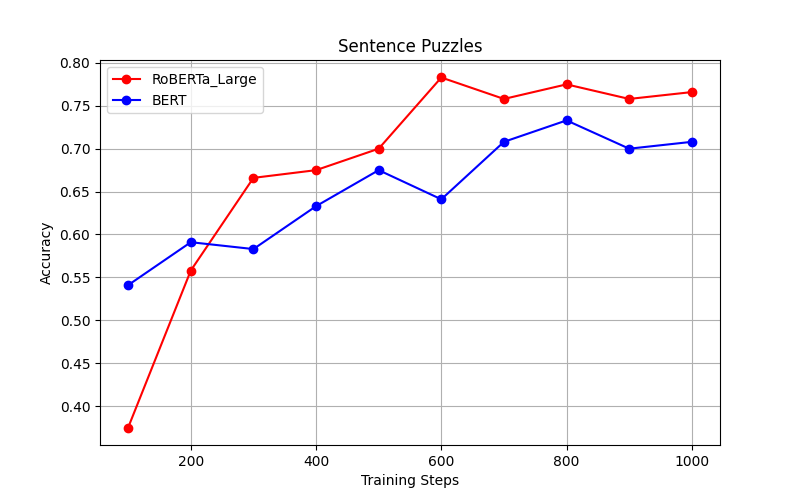}
  \caption{Overall Accuracy of Two Models Logged Every 100 Training Steps on Sentence Puzzles.}
  \label{fig:training_logs_s}
\end{figure}

\begin{figure}[H]
  \centering
  \includegraphics[width=1\linewidth]{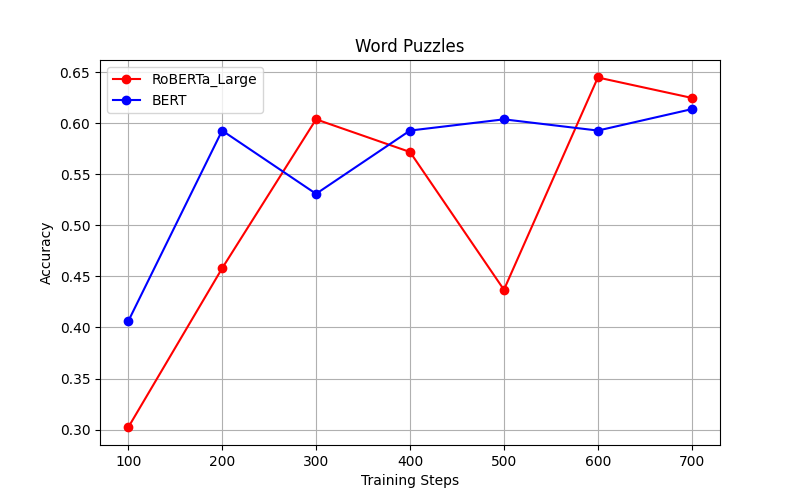}
  \caption{Overall Accuracy of Two Models Logged Every 100 Training Steps on Word Puzzles.}
  \label{fig:training_logs_w}
\end{figure}

\section{Prompts}
\label{sec:prompts}

\begin{figure}[H]
  \centering
  \includegraphics[width=0.8\linewidth]{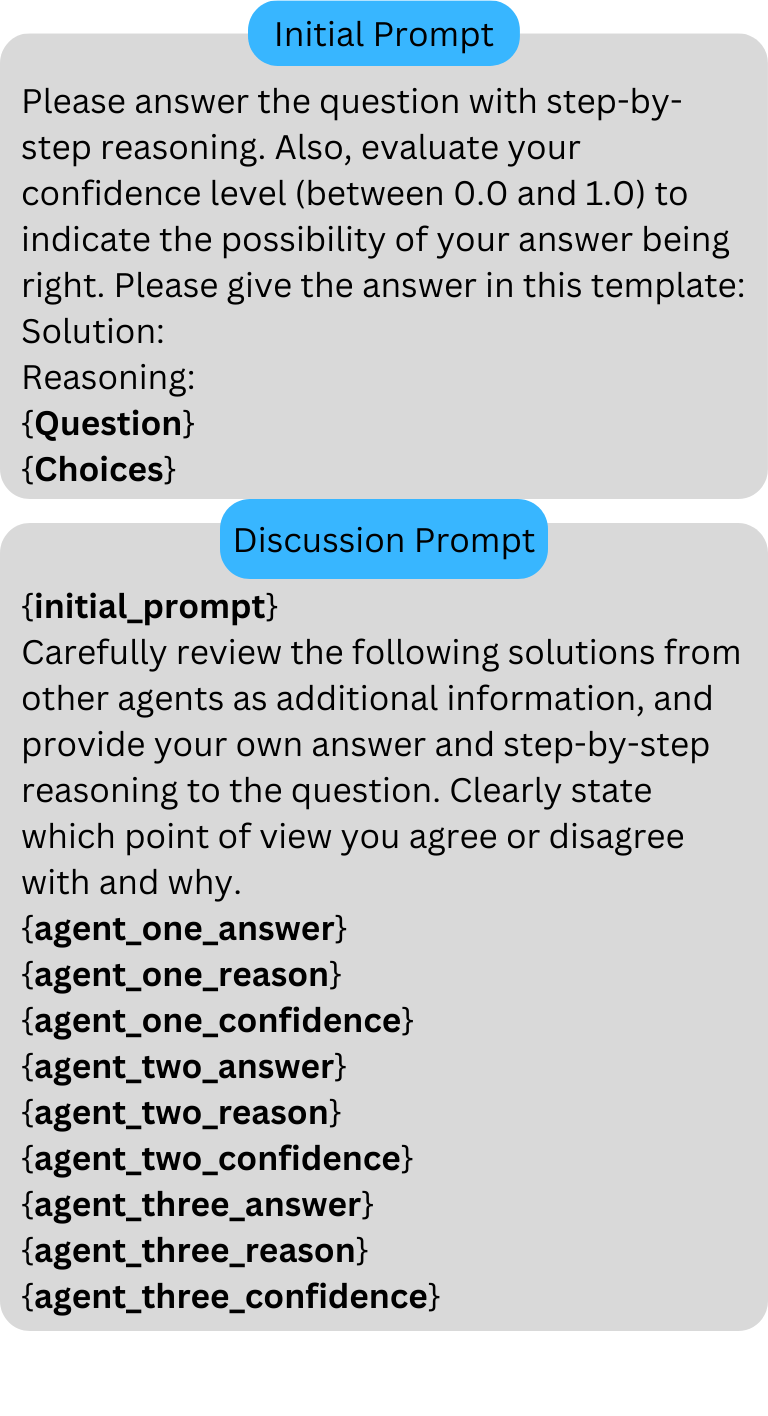}
  \caption{ReConcile Initial and Discussion Prompts}
  \label{fig:prompts}
\end{figure}

\begin{table*}[ht]
\centering
\small
\begin{tabular}{|p{0.45\textwidth}|p{0.45\textwidth}|}
\hline
\multicolumn{1}{|c|}{\textbf{Question}} & \multicolumn{1}{c|}{\textbf{Answer}} \\
\hline
How can you toss a fragile glass ball against a sturdy brick wall from a distance without shattering it? A. No problem, because sturdy bricks are very hard to crack B. You need to throw the glass ball hard.  C. You need to throw the glass ball gently. D. None of above. & Solution: A: No problem, because sturdy bricks are very hard to crack Reasoning: The question is a play on words. It’s not the glass ball that we’re worried about cracking, it’s the brick wall. And brick walls are indeed very hard to crack. \\
\hline
Ana's school has 4 students in Alaska: Maria, James, Jonghyun. What is the name of other student? A. Mark.  B. Ana C. Michael D. None of above.
& Solution: B: Ana. Reasoning: The question states “Ana’s school has 4 students in Alaska: Maria, James, Jonghyun.” So, Ana is the fourth student. \\
\hline
In a house, there are three major rooms. The first is stuffed with cash. The second is stuffed with crucial documents. The third is stuffed with jewels. One day, all of these rooms caught fire. Which room did the cops initially put out the fire? A. The one with jewelry. B. The one with important papers. C. The one filled with money. D. None of above.
 & Solution: D: None of above. Reasoning: The question is a play on words. It’s not the cops’ job to put out fires, it’s the firefighters’ job. \\
\hline
During a 5-day competition, a professional fisherman caught 30 fish. He caught three more fish each day than the day before. On the first day, how many fish did the fisherman catch? A. Two  B. One C. Three D. None of above & Solution: A: Two. Reasoning: If the fisherman caught 30 fish in 5 days, and each day he caught three more fish than the day before, we can set up the equation: x + (x+3) + (x+6) + (x+9) + (x+12) = 30. Solving for x gives us x = 2. So, the fisherman caught two fish on the first day. \\
\hline
\end{tabular}
\caption{Microsoft Copilot Answers to Some Questions}
\label{tab:copilot_examples}
\end{table*}

\section{Copilot Examples}
\label{sec:copilot_examples}

As illustrated in Table \ref{tab:copilot_examples}, the model correctly answers the first three samples but provides an incorrect response for the fourth. This model lacks accuracy in calculation questions, often correctly presenting the equation but solving it inaccurately. Additionally, in some cases, the model accurately formulates the equation and even solves it correctly. However, when the desired option is not among the choices, instead of choosing the 'None of the above' option, it attempts to reinterpret the question in a way that one of the numerical options becomes correct.

\section{Reconcile Convergence}
\label{sec:reconcile_convergence}

Table \ref{tab:reconcile_convergence} indicates the number of questions to which all three models provide the same answer choice in each round. This suggests that after each round, the models achieve a relative convergence.

\begin{table*}[ht]
    \centering
    \begin{tabular}{cc}
        \toprule
        \textbf{Round} & \textbf{Question Count} \\
        \midrule
        Initial Round & 74 \\
        Round 1 & 88 \\
        Round 2 & 104 \\
        \bottomrule
    \end{tabular}
    \caption{ReConcile Models Convergence}
    \label{tab:reconcile_convergence}
\label{tab:reconcile_convergance}
\end{table*}

\section{Reconcile Example}
\label{sec:reconcile_example}
The iterative process of answering the question below is illustrated in Table \ref{tab:reconcile_example}.

\begin{center}
\textbf{If a boat is parked in a marina and tied to a nearby station, how did its owner found the boat a couple of hours later in the middle of the ocean?}\\
A. The station wasn't fixed in the marina\\
B. The dolphins untied the knots.\\
C. The waves were powerful, which pushed the boat and the marina together.\\
D. None of above.
\end{center}

\begin{table*}[ht]
\centering
\small 
\begin{tabular}{|c|c|m{0.45\textwidth}|c|}
\hline
Model & Solution & Reasoning & Confidence \\ \hline
\multicolumn{4}{|c|}{\textbf{Initial Round}} \\ \hline
 GPT3.5 Solution & A & If the station (presumably a mooring or dock) wasn't securely fixed in the marina, it could have drifted into the ocean, taking the boat with it. & 1.0 \\ \hline
 Claude Solution & B & This explains how the boat ended up drifting into the ocean after being tied up. & 0.7 \\ \hline
 Mixtral Solution & A & If the station wasn't fixed properly in the marina, it's possible that the boat became loose and drifted away into the ocean due to wind or tidal influences. Over time, the gap between the boat and the dock could increase, eventually pushing the boat further out into deeper waters. & 0.85 \\ \hline
\multicolumn{4}{|c|}{\textbf{Round 1}} \\ \hline
 GPT3.5 Solution & A  & If the station to which the boat was tied was not properly fixed in the marina, it could have drifted away into the ocean, taking the boat with it. & 0.9 \\ \hline
 Claude Solution & A & I agree with Agents two and three that an unsecured floating station drifting away best explains how the boat ended up adrift in the ocean. & 1.0 \\ \hline
 Mixtral Solution & A & While all options are plausible scenarios for how the boat might have ended up in the ocean, option A provides the most likely explanation given our current understanding of physics and marine environments. Both agent two and agent three suggest that if the station was not fixed properly, it could have been moved by external forces such as wind or tide, thereby pulling the boat along with it. In contrast, option C assumes that the waves were strong enough to move both the marina and the boat simultaneously, an event that seems less probable than the movement of a single object like the station. Option B attributes the boat's displacement to intelligent behavior on part of dolphins, which lack the necessary appendages to manipulate rope knots underwater. Therefore, we can infer that the station must have been unsecured, allowing it to drift out into the ocean while carrying the boat with it. & 0.95 \\ \hline
\multicolumn{4}{|c|}{\textbf{Round 2}} \\ \hline
 GPT3.5 Solution & A & If the station to which the boat was tied was not properly fixed in the marina, it could have drifted away into the ocean, taking the boat with it. & 0.95 \\ \hline
 Claude Solution & A & I agree with Agents two and three that an unsecured floating station drifting away provides the most logical explanation for how the boat ended up adrift.
 & 1.0 \\ \hline
 Mixtral Solution & A & All three agents came to the conclusion that option A is the most likely scenario because it's more feasible that an unsecured floating station would be moved by external forces such as wind or tide, rather than assuming that powerful waves managed to push the marina and the boat together at once, or attributing the boat's displacement to intelligent behavior from dolphins who wouldn't possess the physical capability to manipulate rope knots underwater. & 0.98 \\ \hline
\end{tabular}
\caption{Reconcile Example}
\label{tab:reconcile_example}
\end{table*}

\end{document}